\documentclass[conference]{IEEEtran}
\IEEEoverridecommandlockouts
\usepackage{cite}
\usepackage{amsmath,amssymb,amsfonts}
\usepackage{algpseudocode}
\usepackage[lined,boxed,commentsnumbered,ruled,linesnumbered,longend]{algorithm2e}
\usepackage{graphicx}
\usepackage{textcomp}
\usepackage{xcolor}
\usepackage{subfig}
\usepackage{hyperref}
\hypersetup{
    colorlinks=true,
    linkcolor=blue,
    filecolor=blue,      
    urlcolor=blue,
    citecolor=blue,
}
\usepackage{hyperref}
\usepackage[ruled,linesnumbered]{algorithm2e}
\usepackage[misc]{ifsym} 

\SetKwRepeat{Do}{do}{while}
\def\BibTeX{{\rm B\kern-.05em{\sc i\kern-.025em b}\kern-.08em
    T\kern-.1667em\lower.7ex\hbox{E}\kern-.125emX}}

\DeclareRobustCommand*{\IEEEauthorrefmark}[1]{%
    \raisebox{0pt}[0pt][0pt]{\textsuperscript{\footnotesize\ensuremath{#1}}}}

\begin{document}

\title{VELO: A Vector Database-Assisted Cloud-Edge Collaborative LLM QoS Optimization Framework 
}

\author{
\thanks{This work is supported in part by the National Natural Science Foundation of China (NSFC) under Grant 62272050 and Grant 62302048; in part by Guangdong Key Lab of AI and Multi-modal Data Processing, UIC under Grant 2020KSYS007; in part by Zhuhai Science-Tech Innovation Bureau under Grant 2320004002772; and in part by Interdisciplinary Intelligence Super Computer Center of Beijing Normal University at Zhuhai. \textit{(Corresponding author: Zhiqing Tang.)}}
\IEEEauthorblockN{
Zhi Yao\IEEEauthorrefmark{1,2},
Zhiqing Tang\IEEEauthorrefmark{1}$^{\textrm{\Letter}}$,
Jiong Lou\IEEEauthorrefmark{3}, 
Ping Shen\IEEEauthorrefmark{1}, and
Weijia Jia\IEEEauthorrefmark{1,4}}
\IEEEauthorblockA{\IEEEauthorrefmark{1}Institute of Artificial Intelligence and Future Networks, Beijing Normal University, Zhuhai 519087, China}
\IEEEauthorblockA{\IEEEauthorrefmark{2}School of Artificial Intelligence, Beijing Normal University, Beijing 100875, China}
\IEEEauthorblockA{\IEEEauthorrefmark{3}Department of Computer Science and Engineering, Shanghai Jiao Tong University, Shanghai 200240, China}
\IEEEauthorblockA{\IEEEauthorrefmark{4}Guangdong Key Lab of AI and Multi-Modal Data Processing, BNU-HKBU United International College, Zhuhai 519087, China}
\IEEEauthorblockA{yaozhi@mail.bnu.edu.cn, zhiqingtang@bnu.edu.cn, lj1994@sjtu.edu.cn, iafn@bnu.edu.cn, jiawj@bnu.edu.cn}
}

\maketitle 
\begin{abstract}
The Large Language Model (LLM) has gained significant popularity and is extensively utilized across various domains. Most LLM deployments occur within cloud data centers, where they encounter substantial response delays and incur high costs, thereby impacting the Quality of Services (QoS) at the network edge. Leveraging vector database caching to store LLM request results at the edge can substantially mitigate response delays and cost associated with similar requests, which has been overlooked by previous research. Addressing these gaps, this paper introduces a novel \underline{V}ector database-assisted cloud-\underline{E}dge collaborative \underline{L}LM QoS \underline{O}ptimization (VELO) framework. Firstly, we propose the VELO framework, which ingeniously employs vector database to cache the results of some LLM requests at the edge to reduce the response time of subsequent similar requests. Diverging from direct optimization of the LLM, our VELO framework does not necessitate altering the internal structure of LLM and is broadly applicable to diverse LLMs. Subsequently, building upon the VELO framework, we formulate the QoS optimization problem as a Markov Decision Process (MDP) and devise an algorithm grounded in Multi-Agent Reinforcement Learning (MARL) to decide whether to request the LLM in the cloud or directly return the results from the vector database at the edge. Moreover, to enhance request feature extraction and expedite training, we refine the policy network of MARL and integrate expert demonstrations. Finally, we implement the proposed algorithm within a real edge system. Experimental findings confirm that our VELO framework substantially enhances user satisfaction by concurrently diminishing delay and resource consumption for edge users utilizing LLMs.
	
\end{abstract}

\begin{IEEEkeywords}
	Edge Computing, Quality of Services, Vector Database, Multi-Agent Reinforcement Learning, Large Language Model, Request Scheduling
\end{IEEEkeywords}

\section{Introduction}

The Large Language Model (LLM), as the latest achievement in the field of generative artificial intelligence, 
can be widely used in production and daily life by achieving accurate dialogue service through reasonable prompt text \cite{b1}.
LLM can provide satisfactory answers to users through reasoning, but its extensive parameters demand substantial computational resources, thus prolonging the total time required to generate a comprehensive response for users \cite{b2}.
Additionally, LLMs relying on traditional cloud computing frameworks introduce additional data transfer latency and network traffic stress \cite{9808158}. Conversely, edge computing can offer ample computing power and low latency simultaneously by facilitating collaboration between the edge and the cloud \cite{new_b}.

Current related research primarily focuses on optimizing the challenges of large model sizes and high computational latency by constructing lightweight models \cite{b2}. Techniques such as model quantization and compression are employed directly in the cloud or at the edge to address this issue \cite{new_c1}. This can optimize model parameters by directly reducing the parameter count while minimizing the impact on model performance \cite{new_b3,new_b4}. Other approaches, such as knowledge distillation and model pruning, are utilized to collaboratively fulfill LLM requests based on cloud-edge collaboration between models of different scales \cite{new_b5}.
However, all the mentioned methods involve invasive alterations to the model structure, significantly limiting its versatility. Moreover, all LLM requests still depend on computation, consuming substantial resources, while high latency largely persists.
Therefore, optimizing the Quality of Services (QoS) of LLM with the help of edge servers remains a worthwhile research problem \cite{b6,b7,b71}.

Vector database can cache historical Questions and Answers (QA) as vectors, reducing LLM inference by reusing them when similar requests recur, or enhancing requests through prompt expansion. As a non-invasive LLM optimization technology, it effectively minimizes request completion delay and conserves computational resources while ensuring satisfactory request fulfillment \cite{b8,b9,b10}. 
The main costs of the vector database come from CPU and memory consumption when calculating the similarity between different vectors. As shown in TABLE \ref{tab1}, we have tested the delay required to directly request LLM to return answers and the delay of directly returning answers through database queries at the edge \cite{b15}. Even if the vector database stores $11.34\times 10^6$ vectors, the memory required is only $12.2$ GB. Additionally, the query delay is still very small when the vector database is deployed on the edge server \cite{b11, b12}. Therefore, compared with the large amount of GPU resources and high request delay required to directly deploy LLM on the edge server, deploying a vector database at the edge and storing LLM request results is a very promising method to improve the QoS for edge users.
\begin{table}[t]
	\caption{Comparison between LLM and vector database}
	\label{tab1}
	\begin{center}
		\begin{tabular}{|c|c|c|c|c|}
			\hline
			\textbf{The amount }&\textbf{Loading}&\multicolumn{3}{|c|}{\textbf{LLM request completion delay (s)}} \\
			\cline{3-5} 
			\textbf{of cached } &\textbf{memory}&\multicolumn{2}{|c|}{\textbf{Vector Database}} & {\textbf{Cloud LLM:}}\\
			\cline{3-4} 
			\textbf{vector $(10^{6})$ } &\textbf{(GB)}& \textbf{Edge}& \textbf{Cloud}& \textbf{Qwen14b} \\
			\hline
			$3.64$ &$4$& $0.82$ & $1.05$ & $3.34$ \\
			$9.62$& $10.4$&$0.84$ & $1.08$ & $3.34$ \\
			$10.60$ & $11.4$&$0.83$ & $1.03$ & $3.34$ \\
			$11.34$ & $12.2$&$0.81$ & $1.05$ & $3.34$ \\
			\hline
		\end{tabular}
	\end{center}
\end{table}

We propose a novel \underline{V}ector database-assisted cloud-\underline{E}dge collaborative \underline{L}LM QoS \underline{O}ptimization (VELO) framework. In the VELO framework, we deploy the vector database on edge servers and cache some results returned by the LLM. The scheduling decisions of new LLM requests are made based on requests features and vector similarity in the edge vector database. Specifically, the user first offloads the LLM request to the nearest edge server. Then, the edge server chooses one of the following processing methods to return the answer for this request: 1) Query the request directly from the edge vector database and return the answer. 2) Utilize similar vectors in the edge vector database to enhance the user request, and then request the LLM from the cloud to return the answer. 3) Directly request the LLM from the cloud to return the answer.

However, several challenges remain unresolved when determining whether LLM requests should be processed by the edge or the cloud. Firstly, the correlation between LLM requests is significant, and there are new features in this scenario, such as one question for multiple answers, multiple queries corresponding to one answer, and timeliness of request results \cite{b14, b16}. Secondly, traditional scheduling methods base decisions on analyzing the similarity between newly arrived requests and the cached vectors. However, different LLM requests exhibit varying sensitivities to the similarity between requests, reflected in the diverse forms and descriptions of language requests \cite{b17}. Additionally, the features of LLM requests are discrete, which can pose challenges in early exploration and lead to data wastage in related training models \cite{x3}. Fortunately, Reinforcement Learning (RL) can effectively and dynamically consider the relationship between complex LLM requests and vectors stored in the vector database through the learning of policy networks and the design of rewards, thereby making LLM request scheduling decisions with higher long-term returns \cite{b18}.

We present a distributed \underline{L}LM \underline{R}equest \underline{S}cheduling (LRS) algorithm utilizing Multi-Agent Reinforcement Learning (MARL) to enhance the scheduling process of LLM requests \cite{b18, x4}. The RL agent is placed on each edge server to determine LLM request scheduling. First, to address the challenges of feature extraction and similarity analysis of diverse requests, we introduce a request feature extraction network built on the Transformer Encoder \cite{b25,9954278}. This network merges various request features with vector query outcomes from the edge vector database, thereby boosting the learning capacity of the policy network. By employing Centralized Training and Decentralized Execution (CTDE), the edge vector database can offer users high-quality, low-delay services efficiently \cite{b23,x5}. Secondly, to address the discrete nature of LLM requests, we suggest a policy network training approach based on expert demonstrations \cite{b19}. The network is updated with the support of similar decision-making agents to achieve superior performance. The integration of expert demonstrations effectively resolves concerns regarding poor vector richness and challenges in model fitting due to early action sampling.


In this paper, we propose the VELO framework to optimize the QoS of LLM at the network edge by deploying vector databases at edge servers. Additionally, we design the LRS algorithm to determine whether a user's LLM request should be processed in the cloud or at the edge. We have made numerous enhancements to the algorithm, such as incorporating feature extraction modules and expert demonstrations. To validate the effectiveness of the VELO framework, we deploy a real edge system comprising a cloud and multiple edge servers. The open-source model Qwen is deployed and utilized as an LLM in the cloud \cite{b22}, while public datasets serve as LLM requests from users \cite{b20, b31}. Experimental results demonstrate that our VELO framework and LRS algorithm can effectively enhance the QoS of LLMs at the edge. The main contributions of this paper are summarized as follows:

\begin{itemize}
	\item We introduce the vector database-assisted cloud-edge collaborative LLM QoS optimization framework, VELO. In VELO, vector databases are deployed on edge servers to store LLM request processing results. This framework is highly versatile, maintaining the structure of LLMs and applicable across various LLM implementations.
	\item  We propose the LRS algorithm based on MARL to determine whether an request should be processed in the cloud LLM or on an edge server. Additionally, to enhance feature extraction and convergence performance, we incorporate a feature extraction network and include expert demonstrations during training.
	\item  We have implemented the VELO framework and LRS algorithm in a real edge system, complemented by larger-scale simulations using virtual machines. Experimental results indicate the efficacy of our algorithms in enhancing the QoS of edge users when requesting LLMs, leading to higher satisfaction and lower latency.
\end{itemize}

The remainder of the paper is organized as follows. In Section \ref{sec2}, the VELO framework and problem formulation are described. The LRS algorithm is proposed in Section \ref{sec3}. The system implementation and experimental results are described in Section \ref{sec4}. Finally, Section \ref{sec5} concludes the paper.

\section{VELO framework and Problem Formulation}
\label{sec2}

\subsection{VELO Framework}\label{AA}

\begin{figure}[!h]
	\centering
	\includegraphics[width=1\linewidth]{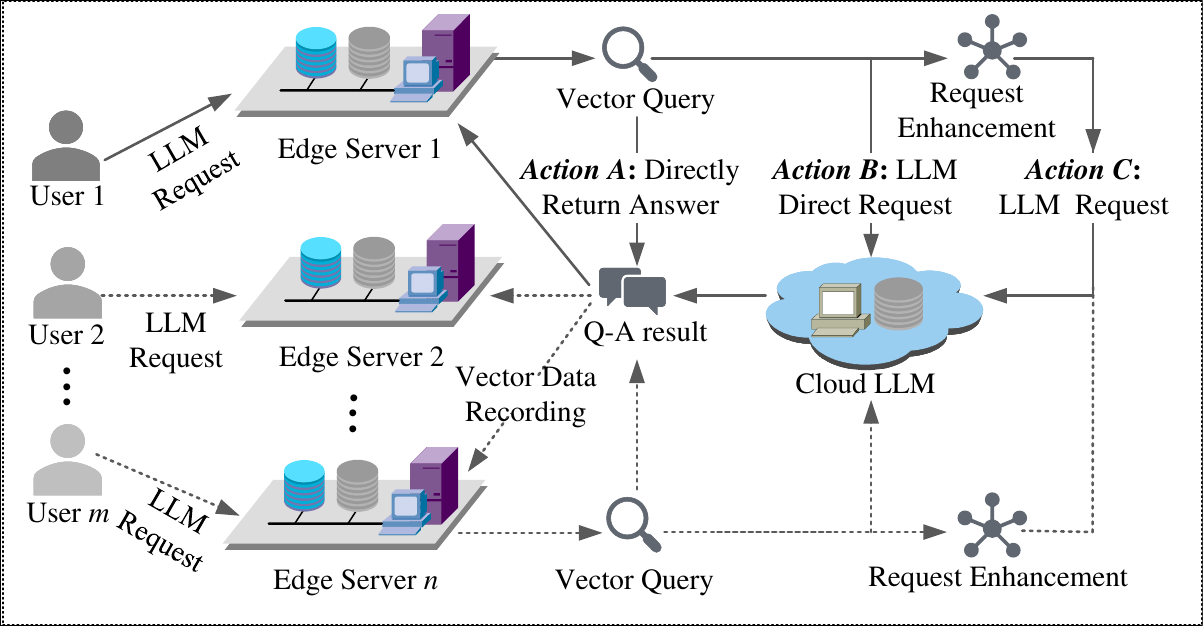}
	\centering
	\caption{VELO framework overview.}
	\label{fig:system model}
\end{figure}

The VELO framework, illustrated in Fig. \ref{fig:system model}, comprises users, edge servers, vector databases, and the cloud LLM. Various users dispatch distinct LLM requests to nearby edge servers. Upon receiving a user request, the edge server decides how to handle it based on the request's content and subsequently provides the result. Using Edge Server 1 as an illustration, upon receiving a request from User 1, it initially queries the local vector database and then assesses request features alongside vector query outcomes. Based on this analysis, it selects LLM request scheduling actions from the following options: \textit{Action A} - returning the answer directly from the vector database, \textit{Action B} - directly requesting the cloud LLM and returning the answer, and \textit{Action C} - augmenting the user request with the vector database query results and requesting the LLM in the cloud to return the answer. Consequently, edge servers can offer high QoS request scheduling decisions through cloud-edge collaboration, predominantly assessed by completion satisfaction and delay. The specifics are elaborated as follows.

\textbf{Users and edge servers:} There exists a set of mobile users $\mathbf{M}$ and a set of edge servers $\mathbf{N}$. At each time slot $t$, the LLM request is generated by user $m \in \mathbf{M}$ and offloaded to edge server $n \in \mathbf{N}$, which can be represented as $x_{m,n}(t)$. Subsequently, the LLM request is embedded as a vector $\boldsymbol{f}_{m}(t)$. The experience samples generated by the server processing the LLM request is $E_{n}$, with a quantity of $l_{\text{n}}$. The experience generated by all agents can be represented as $\mathbf{E}_{N}$. 
In addition, there are expert demonstrations $E_{\text{g}}$, with a quantity of $l_{\text{g}}$, cached in advance on the server.

\textbf{Vector database:} Each edge server $n$ has a vector database $\mathbf{V}_{n}(t)$ to cache request completion records, with a current data volume of $k(t)$ \cite{b23}.
A vector query result of an LLM request in $\mathbf{V}_{n}(t)$ is a vector data collection $\mathbf{L}_{m,n}(t)$ of $P$ items with the highest similarity to the request. 
The $p$-th item of data in $\mathbf{L}_{m,n}(t)$ can be described as $\boldsymbol{l}_{m,n,p}(t)$. Vector cache value $c_{n,p}^{\text{r}}(t)$ is cached in the $\mathbf{V}_{n}(t)$, with an average value of $\bar{c_{n}^{\text{r}}}(t)$. 

The correlation between request $\boldsymbol{f}_{m}(t)$ and vector query data $\mathbf{L}_{m,n}(t)$ can be expressed as $\boldsymbol{c}_{m,n}(t)$. Then, the correlation between $\boldsymbol{f}_{m}(t)$ and one of the vector query data $\boldsymbol{l}_{m,n,p}(t)$ can be expressed as $c_{m,n,p}(t)=\{c_{m,n,p}^{\text{s}}(t),c_{m,n,p}^{\text{k}}(t),c_{m,n,p}^{\text{f}}(t)\}$. In which, $c_{m,n,p}^{\text{s}}(t)$ denotes the value of similarity between $\boldsymbol{f}_{m}(t)$ and $\boldsymbol{l}_{m,n,p}(t)$, $c_{m,n,p}^{\text{f}}(t)$ is the number of times vector data $\boldsymbol{l}_{m,n,p}(t)$ has been used, and $c_{m,n,p}^{\text{k}}(t) \in \{1,2\}$ denotes the type of $\boldsymbol{l}_{m,n,p}(t)$ including question and answer. 
\subsection{QoS Definition}

As shown in Fig. \ref{fig:system model}, edge servers provide high QoS request scheduling decisions through cloud-edge collaboration, which is mainly measured by request completion satisfaction $q_{m,n}(t)$ and request completion delay $d_{m,n}(t)$ \cite{bqos}. 

\textbf{Request completion satisfaction:} The similarity between LLM requests and vector data $\boldsymbol{l}_{m,n,p}(t)$ can be calculated based on L2 Euclidean distance \cite{b15}.
\begin{equation}
	\label{distance}
 	J_{m,n,p}(t) = \sqrt{\sum_{h=1}^{H}\left(  f_{m,h}(t)- l_{m,n,p,h}(t) \right )^{2} }.
\end{equation}
where the dimension of $\boldsymbol{f}_{m}(t)$ and the vector data obtained from query $\boldsymbol{l}_{m,n,p}(t)$ is $H$. The $q_{m,n}(t)$ is used to evaluate the satisfaction of completing LLM request at time $t$. When the reference answer of the LLM request is known, $q_{m,n}(t)$ is measured by the similarity between the current answer and the reference answer, which is calculated as  Eq. (\ref{qmn}).
\begin{equation}
	\label{qmn}
	q_{m,n}(t) = - \sqrt{\sum_{h=1}^{H}\left(  f_{m,h}^{\text{a}}(t)- f_{m,h}^{\text{r}}(t) \right )^{2} }.
\end{equation}
where $f_{m}^{\text{a}}(t)$ and $f_{m}^{\text{r}}(t)$ are the answers  obtained through LRS algorithm and reference answers. 

\textbf{Request completion delay:} The $d_{m,n}(t)$ represents the delay in the return of request $x_{m,n}(t)$, which is determined as follows: 
\begin{equation}
\label{formula-delay}
	d_{m,n}(t)=\begin{cases}
		& d_{m,n}^{\text{e}}(t), \quad \textit{Action A}\\  
		& d_{m,n}^{\text{c}}(t), \quad \textit{Action B} \\ 
		& d_{m,n}^{\text{e}}(t) + d_{m,n}^{\text{c}}(t), \quad \textit{Action C}.\\ 
	\end{cases}
\end{equation}
In Eq. (\ref{formula-delay}), $d_{m,n}^{\text{e}}(t)$ is the delay for the system to complete LLM requests through \textit{Action A},
$d_{m,n}^{\text{c}}(t)$ is the delay for the system to complete LLM requests through \textit{Action B}, 
and $d_{m,n}^{\text{e}}(t) + d_{m,n}^{\text{c}}(t)$ is the delay for the system to complete LLM requests through \textit{Action C} including the time when the server acquires cache knowledge and the time when the LLM is requested.

\subsection{Vector Database Operations}
When the LLM request is resolved by \textit{Action B}, the server embeds the QA and inserts their information including vector embedding, $c_{m,n,p}^{\text{k}}(t)$, $c_{m,n,p}^{\text{f}}(t)$, and vector cache value $c_{n,p}^{\text{r}}(t)$ separately into the vector database. 
When the server processes LLM requests through \textit{Action A} and \textit{Action C}, vector data most relevant to the LLM request $\boldsymbol{l}_{m,n,p}(t)$ in the request query result $\mathbf{L}_{m,n}(t)$ is filtered to assist in LLM request processing. The principles of filtering can be represented as $
{ \underset {p\in P} { \operatorname {arg\,max} } \, F_{n, p}(t)} 
$, which can be further expressed as \cite{b13}:
\begin{equation}
	\label{selection_item}
	\begin{split}
		F_{n, p}(t) = \phi_{1} c_{m,n,p}^{\text{s}}(t) + \phi_{2}c_{m,n,p}^{\text{f}}(t), p \in P.
	\end{split}
\end{equation}
where $\phi_{1}$ and $\phi_{2}$ are weights used to balance the effective of these factors. In addition, the cache value of query vector is updated as:
\begin{equation}
	c_{n,p}^{\text{r}}(t)=(c_{n,p}^{\text{r}}(t-1)+q_{m,n}(t)-d_{m,n}(t))/2 .
\end{equation}
which is beneficial for analyzing and managing cached vector data. When the cached data is correctly matched to the request, $c_{n,p}^{\text{r}}(t)$ will increase, and vice versa. 

\subsection{Problem Formulation}
We aim to maximize the QoS of the LLM request for the system, which mainly depends on Eq. (\ref{qmn}) and Eq. (\ref{formula-delay}). The goal is to find the best policy to enhance QoS while adhering to constraints. The definition of LRS problem is as follows:

\begin{align*}
&\min W(t)=\sum _{m\in M} \sum _{n\in N}  \big(-\varphi _{1} q_{m,n}(t) + \varphi _{2}d_{m,n}(t) \big)\\
& \begin{array}{r@{\quad}r@{}l@{\quad}l}
s.t.&q_{m,n}(t)&< 0,  \forall m \in M, \forall n \in N,\\
&d_{m,n}(t)&>0,  \forall m \in M, \forall n \in N,\\
&c_{m,n,p}^{\text{k}}(t)&\in \{1,2\},  \forall m \in M, \forall n \in N,  \forall p \in P,\\
&c_{m,n,p}^{\text{r}}(t)& < 0,  \forall m \in M, \forall n \in N,  \forall p \in P,\\
&c_{m,n,p}^{\text{s}}(t)& > 0,  \forall m \in M, \forall n \in N,  \forall p \in P,\\
&c_{m,n,p}^{\text{f}}(t)& > 0,  \forall m \in M, \forall n \in N,  \forall p \in P\\
\end{array} 
\end{align*}

The LRS problem is NP-hard and can only be solved heuristically. 
However, most heuristic algorithms make scheduling decisions based on deterministic policies and cannot consider the effects of dynamic environments and continuous decisions. For meta-heuristic algorithms, it is necessary to know all future information, but the future LLM requests are unknown.
The Greedy strategy, a common strategy for heuristic algorithms, makes judgements based on the similarity between the question and the database content, but different types of requests have different sensitivities to similarity. Whereas the arrival of LLM requests and updates to the environment are memoryless, so this problem can be modeled as a Markov Decision Process (MDP) \cite{b24}.

To solve the MDP problem, RL is a promosing method and has been widely adopted. By treating each server as an agent, we propose our LRS algorithm based on Multi-Agent Proximal Policy Optimization (MAPPO) to make the LLM request scheduling decisions \cite{x6}. 
Through LRS training, the agent considers the associated status of cached vector databases and then selects the action from a global perspective.
The long-term QoS of the system for handling LLM requests can be improved by the reward function.

\section{Our Algorithms}
\label{sec3}
\subsection{ Algorithm Settings}

The LRS algorithm is founded on MAPPO, with an agent deployed on each edge server to make scheduling decisions independently. Each agent maintains a local state and shares a policy. Furthermore, a global value function incorporates global information and updates the policy network, enabling multiple agents to collaborate and optimize for better long-term benefits for the system. The main settings are outlined as follows.

\textbf{State:} The state provides a comprehensive description of LLM request and the queried vectors. It encompasses the correlation between the LLM request and the vectors, as well as the features of the request. Therefore, the state of the agent on edge server $n$ at time slot $t$ can be divided as follows.

\textit{Correlation State}: The correlation information between the LLM request and the edge vector database is represented by a matrix $\boldsymbol{c}_{m,n}(t)$, which can be denoted as:
\begin{equation}
\boldsymbol{c}_{m,n}(t)=\begin{bmatrix}
	c_{m,n,1}^{\text{s}}(t) & \ldots & c_{m,n,p}^{\text{s}}(t) \\ 
	c_{m,n,1}^{\text{k}}(t) & \ldots & c_{m,n,p}^{\text{k}}(t) \\ 
	c_{m,n,1}^{\text{f}}(t) & \ldots & c_{m,n,p}^{\text{f}}(t) 
\end{bmatrix}.
\end{equation}

\textit{Request Feature State:} Considering the different sensitivity of various requests to the vector database, it is crucial to include the request features in the state representation. To accomplish this, we employ a request embedding tool based on the Transformer Encoder \cite{b28}, defined as:
\begin{equation}
	\label{formula-13}
	\boldsymbol{f'}_{m}(t)= T[\boldsymbol{f}_{m}(t)],
\end{equation}
where $T[\cdot]$ represents the network layers based on the Transformer Encoder and fully connected layer used to extract the features of the initial request vector $\boldsymbol{f}_{m}(t)$. Then, the local state of each agent is obtained as:
\begin{equation}
	\label{s_mnt}
	\boldsymbol{s}_{n}(t)= \left \{  \boldsymbol{c}_{m,n}(t), \boldsymbol{f'}_{m}(t) \right \}.
\end{equation}
And the global state can be denoted as:
\begin{equation}
        \label{s_t}
	\boldsymbol{s}(t)= \left \{  \boldsymbol{s}_{n}(t)|m \in \mathbf{M}, n \in \mathbf{N} \right \}.
\end{equation}

\textbf{Action:} Each action refers to the scheduling of the LLM request, which can be denoted as:
\begin{equation}
	a_{n}(t) \in \left \{  0,1 \right \},
\end{equation}
where $a_{n}(t)=0$ indicates that the LLM request is processed by the edge vector database, with \textit{two sub-actions}: either returning the results directly from the vector database (\textit{Action A}) or using the vector database to enhance the request before querying the LLM in the cloud (\textit{Action C}). Conversely, $a_{n}(t)=1$ signifies that the request is directly forwarded to the LLM in the cloud (\textit{Action B}).

\textbf{Reward:} The objective of each agent is to maximize the reward. In the VELO framework, the aim is to enhance the QoS, encompassing increasing user satisfaction and reducing request completion delay, which can be denoted as:
\begin{equation}
\label{formula-rt}
         r_{n}(t) = -W(t).
\end{equation}

\begin{figure*}[ht]
	\centering
	\includegraphics[width=7in]{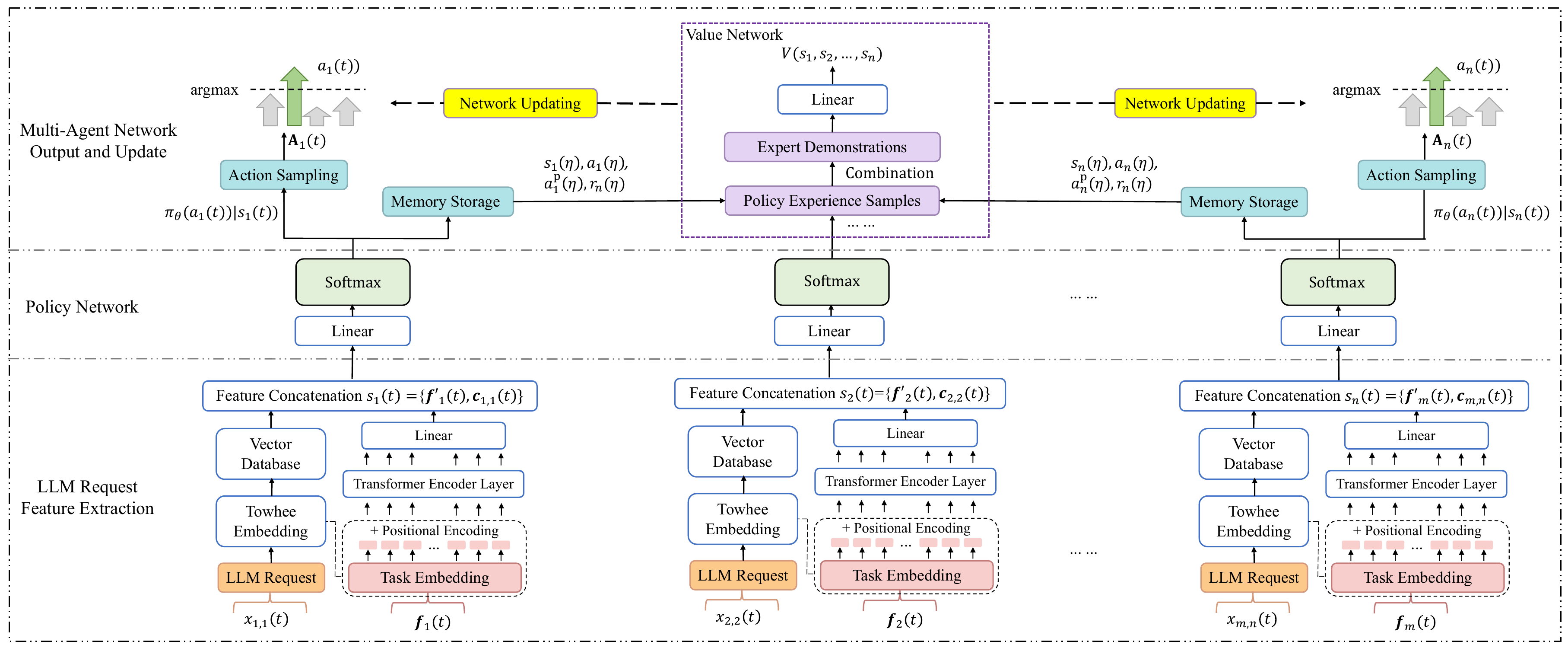}
	\centering
	\caption{LRS Algorithm Overview.}
	\label{fig:model structure}
\end{figure*}

\textbf{Policy:} The policy is utilized to select actions, typically represented by $\pi$ \cite{b26}. Each agent selects an action based on local states and a shared policy $\pi_\theta(a_{n}(t)|\boldsymbol{s}_{n}(t))$ to maximize the cumulative discounted reward $J_{\theta}$, which is denoted as:
\begin{equation}
	J_{\theta}=\mathbb{E}_{\boldsymbol{s}_{n}(t),a_{n}(t)}\left [ \Sigma _{t}\gamma r(t) \right ].
\end{equation}
where $\gamma$ is a discount factor.

\subsection{Vector Database-assisted Feature Extraction}

The LLM Request Feature Extraction module is described in Fig. \ref{fig:model structure}. By extracting the LLM request query result features from the vector database and combining the LLM request features, the high-dimensional LLM request vectors are mapped while the LLM information is fully extracted.
As shown in Fig. \ref{fig:model structure}, each edge server, acting as an agent, independently performs the request feature learning process.

Taking agent $n$ as an example, we analyze the process from bottom to top. LLM request is first encoded into vector $\boldsymbol{f}_{m}(t)$ through the \textit{Towhee} framework and then utilized through two pathways. 
The LRS performs a vector query in the vector database to obtain the matching results between LLM and vector database content after comparing $\boldsymbol{f}_{m}(t)$ with vector data $\mathbf{L}_{m,n}(t)$. 
In addition, LRS feeds $\boldsymbol{f}_{m}(t)$ to the Transformer Encoder, which consists of a position encoder and $z$ multi-head attention modules to obtain a better representation of the task features $\boldsymbol{f}_{m}^{\text{e}}(t)$. Then $\boldsymbol{f}_{m}^{\text{e}}(t)$ is further compressed through a linear layer to obtain the LLM request features $ \boldsymbol{f'}_{m}(t)$ for the inputs of policy network and value network. Finally, features $\boldsymbol{c}_{m, n}(t)$ and $\boldsymbol{f'}_{m}(t)$ will be connected.

	

\subsection{Training with Expert Demonstrations}

To solve the problems of sparse training data and slow convergence at the early stage of training, we add expert demonstrations during the training process, as shown in Algorithm \ref{algorithm:Alg2}. 
To ensure that each network update is valuable, we set the minimum number of expert demonstrations and experience required for network training and updating to $l_{\min}^{\text{g}}$ and $l_{\min}^{\text{m}}$, respectively.
Then, we denote the number of network updates as $u(u<u_{\max})$.

Algorithm \ref{algorithm:Alg2} is deployed on the edge server for network training. Through LRS training, the weights of the network $b(t)$ are updated and sent down to all agents. Algorithm \ref{algorithm:Alg2} describes the expert demonstrations assisted network training and updating process in the Multi-Agent Network Policy and Update module in Fig. \ref{fig:model structure}.
As shown in Fig. \ref{fig:model structure}, after processing LLM requests, edge servers cache the experience locally and send it periodically to the server for network training. If the server has collected enough experience for each agent, it will train the network according to lines 7 - 21. The number of expert demonstration items $l_{\text{g}}^{\text{u}}$ used in training decreases as the number of network updates increases. 
When $l_{\text{g}}^{\text{u}}$ is above the threshold $l_{\min}^{\text{g}}$, the server will sample the expert demonstrations, train and update the network according to lines 10 - 12. Conversely, considering that the expert demonstrations are outdated for the network, it will be trained and updated using only $\mathbf{E}_{N}$ according to lines 13 - 14.


\textbf{Policy optimization:}
We use the policy gradient method to update the network parameters. The policy estimation of time step $t$ for parameter $\theta$ can be calculated as 
\begin{equation}
\hat{g}_t(\theta) = \frac{1}{N} \sum_{n=1}^{N} \mathbb{E}_{\tau_n} \left[ \sum_{t=0}^{T_n} \nabla_\theta \log \pi_\theta(a_{n}(t)|\boldsymbol{s}_{n}(t)) \hat{A_{n}}(t)\right].
\end{equation}
where $\mathbb{E}_{\tau_n}$ represents the expectation for the trajectory of agent $n$, $\pi_\theta \left ( a_{n}(t)|\boldsymbol{s}_{n}(t) \right )$ is the policy function of agent $n$, and $\hat{A_{n}}(t)$ is an adjusted advantage function introduced in the LRS algorithm to handle the mutual influence between agents in multi-agent environments, which can be denoted as:
\begin{equation}
\label{advantages}
\hat{A}(t) = \delta (t) + \gamma \lambda\delta (t+1)+ \ldots + (\lambda\delta)^{T-t+1}\lambda (T-1).
\end{equation}
where $\delta (t) = r(t)+ \gamma V^{\pi}\left ( \boldsymbol{s}(t+1) \right ) - V^{\pi}\left ( \boldsymbol{s}(t) \right )$, and if the agent starts in state $s(t)$ and takes action according to policy $\pi$, the value function $V^{\pi}\left ( \boldsymbol{s}(t) \right )$ gives the expected return, which can be denoted as:
\begin{equation}
	V^{\pi}\left ( \boldsymbol{s}(t) \right )=\mathbb{E}_{\tau \sim\pi}[r(\tau)|\boldsymbol{s}=\boldsymbol{s}(t)].
\end{equation}
The loss function of the policy gradient is:
\begin{equation}
	L_{\text{PG}}(\theta) = \hat{\mathbb{E}}[log\pi_{\theta}(a(t)|\boldsymbol{s}(t))\hat{A}(t)].
\end{equation}
Moreover, the loss function is constrained to ensure that the difference between new and old parameters is not too large, which is further expressed as: \cite{b27}:
\begin{equation}
\label{loss}
	L(\theta) = \hat{\mathbb{E}}[L_{\text{clip}}(\theta) -c_{1}L_{\text{E}}(\theta)+c_{2}C_{\text{e}}[\pi_{\theta}](\boldsymbol{s}(t))].
\end{equation}
where $L_{\text{E}}(\theta)$ is the mean square error of the reward and the corresponding state value, $C_{\text{e}}[\pi_{\theta}](\boldsymbol{s}(t))$ is the cross entropy of the action probability distribution, $c_{1}$ and $c_{2}$ are hyperparameters, with values of -0.5 and 0.01, respectively. The $L_{\text{clip}}(\theta)$ is denoted as:
\begin{equation}
	L_{\text{clip}}(\theta) = \hat{\mathbb{E}}[\min(\psi _{t}(\theta)\hat{A}(t),\text{clip}(\psi _{t}(\theta),1-\epsilon,1+\epsilon )\hat{A}(t))].
\end{equation}
where $\psi_{t}(\theta)$ is the probability ratio of actions under two policies$, \text{clip}(\psi_{t}(\theta),1-\epsilon,1+\epsilon )$ is used to clip the $\psi_{t}(\theta)$ between $(1-\epsilon, 1+\epsilon)$, and
$\epsilon$ is a hyper-parameter, e.g., $\epsilon=0.2$, which can make the value of actions lower or higher than the average amplitude between $(1-\epsilon, 1+\epsilon)$. 

\begin{algorithm}[t]
	\caption{LRS Training}
	\label{algorithm:Alg2}
	\textbf{Input:} $E_{\text{g}}$, $\mathbf{E}_{N}$, $u$, $l_{\text{g}}$, $l_{n}$  \\
        \textbf{Output:} $b(t)$\\
        Initialize policy network $\pi_\theta$;\\
	Initialize $u=0$, $l_{n}=0$;\\
	\While{$u<u_{\max}$}{
            Check $\mathbf{E}_{N}$ and calculate $l_{n}, n=(1,\ldots, N)$;\\
		\If {$\forall l_{n}>l_{\min}^{\text{m}}, n=(1,\ldots, N)$} {
  	    $l_{\text{g}}^{\text{u}} =              l_{\text{g}} / u $;\\
			\For{$n = 1,2, \ldots ,N$}{
				\uIf {$l_{\text{g}}^{\text{u}} > l_{\min}^{\text{g}}$} {
					Sample $l_{\text{g}}^{\text{u}}$ expert demonstrations from $E_{\text{g}}$ to get $E_{g}'$;\\
					Get training experience $\{ E_{g}',\mathbf{E}_{N} \}$;\\
				} \Else {
					Get training experience $\mathbf{E}_{N} $;\\
				}
                Compute $\hat{A}(1), \ldots ,\hat{A}(T)$ by Eq. (\ref{advantages});\\
                Compute $L(\theta)$ by Eq. (\ref{loss});\\
			Optimize the network and get $b(t)$;\\
			$u =u+1$;\\
                Update weights $\pi_{\text{old}} \leftarrow \pi_{\theta}$;
			}
            Send $b(t)$ to all agents;\\
            Drop $\mathbf{E}_{N}$ on the edge server;\\
		} 
	}
\end{algorithm}

\subsection{LLM Request Scheduling}
Algorithm \ref{algorithm:Alg3} describes the LRS algorithm, whose output is the reward value $r_n(t)$ that represents the QoS of agent $n$. The update time for local network weights is $b_{n}^{\text{T}}(t)$ and the periodicity of vector database content checking and updating is $\xi$. 
First, agent $n$ creates a collection of vector databases and obtains the update time of the local network weights, as shown in lines 5 - 6. 
Then, the server processes the arriving LLM requests and stores experience, as shown in lines 12 - 14. 
In each time slot, the LRS inputs $\boldsymbol{s}_{n}(t)$ into the policy network to obtain the corresponding action $a_{n}(t)$ and the probability of action execution $a_{n}^{\text{p}}(t)$. Then the agent $n$ samples actions based on the $a_{n}^{\text{p}}(t)$. In addition, the server periodically executes the vector data eviction policy, as shown in lines 15 - 22. 

When the experience accumulated by agents is sufficient, it is sent to the global value network, as shown in lines 23 - 26. Then, after completing the network update, new weights are sent to all agents. 

\textbf{Vector data eviction policy:}
Low-quality data records resulting from poor policies need to be updated and processed promptly. 
Therefore, in addition to data insertion and update operations, we also introduces checks and updates for vector data quality. 
For each elapsed time $ \xi $, 
data in the vector database will be dropped when they satisfy $c_{n,p}^{\text{r}}(t) < \bar{c_{n}^{\text{r}}}(t)$, as these data usually exhibit characteristics such as training failures, mismatches, and outdated answers.

\begin{algorithm}
	\caption{The LRS Algorithm}
	\label{algorithm:Alg3}
	\textbf{Input:} $T$, $b_{n}^{\text{T}}(t)$, $u$, $x_{m,n}(t)$, $l_{n}$, $\xi$, $c_{n,p}^{\text{r}}(t)$ $k(t)$\\
        \textbf{Output:} $r_{n}(t)$\\
	Initialize $u=0,t=0$;\\
	Create vector data collection;\\
	  $b_{n}^{\text{i}}=b_{n}^{\text{T}}(0)$;\\
	Reset environment and get $s_{n}(0)$ ;\\
	
		\While{$u<u_{\max}$ and $t<T$}{
   			\If {$b_{n}^{\text{T}}(t) \neq b_{n}^{\text{i}}$} {
				Update networks by calling Algorithm \ref{algorithm:Alg2}; \\
                $b_{n}^{\text{i}} = b_{n}^{\text{T}}(t)$; \\
			}
			Get $a_{n}(t),a_{n}^{\text{p}}(t)$ by $s_{n}(t)$;\\
			Execute $a_{n}(t)$ and get $r_{n}(t)$;\\
			Store $\{s_{n}(t)$, $a_{n}(t)$, $a_{n}^{\text{p}}(t)$, $r_{n}(t) \} $ in $l_{n}$;\\
			\If (\tcp*[h]{Vector data eviction policy}){$t \% \xi = 0 $} {
			$ \bar{c_{n}^{\text{r}}}(t) = \frac{1}{k(t)}\sum_{p=1}^{k(t)} c_{n,p}^{\text{r}}(t) $; \\
                   \For{$p = 1,2, \ldots, k(t) $}{
                    \If {$c_{n,p}^{\text{r}}(t) < \bar{c_{n}^{\text{r}}}(t)$} {
			Drop the vector data; \\}
			}
                } 
                \If {$t \% l_{\min}^{\text{m}} = 0 $} {
			Send experience to the training server; \\
                $l_{n}=0$\\}
                Get $s_{n}(t+1)$ based on $x_{m,n}(t+1)$;  \\
                $s_{n}(t)  = s_{n}(t+1)$; \\
                $u =u+1$;\\
                $t = t +1$;\\
			 
		} 
\end{algorithm}

\subsection{Computational Complexity Analysis}
The LRS algorithm can be mainly divided into four parts: state observation, action selection, reward calculation, and network updates. The computational complexity of each section is analyzed below.
First, the state is shown in Eq. (\ref{s_t}), and the complexity of this part can be calculated to be $O(|\textbf{N}|)$, where $|\textbf{N}|$ represent the number of servers. Secondly, the complexity of action selection for all servers can be expressed as $O(|\textbf{N}|)$. The complexity of reward calculation does not vary with the number of servers, so its complexity can be expressed as $O(1)$. The extracted requests features are mapped through fully connected layers, with $L_{1}$ hidden layers and $G$ neurons in each layer. The complexity of this section can be calculated as $O (| \textbf{N} | \times G+L_{1} \times G^{2})$ \cite{b33}. 

The computational complexity of the Transformer Encoder module is primarily influenced by its multi-head self-attention operation, which is one of its key components. This complexity is determined by the embedding dimension $H$ and the number of patches $C$, which can be expressed as $O(H^{2}C + HC^{2})$ \cite{b34}.
Assuming that the Transformer Encoder module contains $L_{2}$ layers, its overall complexity can be calculated as $O(L_{2} \times (H^{2}C + HC^{2}))$.

The total amount of experience used for each update is fixed for network updates, and the ratio of expert experience to the experience of agents varies. The other operations in the LRS have a relatively small impact on computational complexity analysis. Therefore, the complexity of the LRS algorithm is $O (| \textbf{N} | \times G+L_{1} \times G^{2} + L_{2} \times (H^{2}C + HC^{2}))$. 

\section{System Implementation and Experiments}
\label{sec4}



\subsection{System Implementation}
We have implemented a VELO prototype system and deployed the LRS algorithm into the system using Python. The VELO system consists of edge servers and cloud LLM, where the edge mainly consists of \textit{Towhee} Service and \textit{Milvus} Service \cite{b15}, \cite{b29}, all of which are deployed through containers. 
The experimental platform consists of three desktops featuring an Intel i9-10900K 10-Core CPU and NVIDIA RTX 2070 Super GPU. Data from edges is transmitted to a desktop for network training, which is equipped with an Intel i7-13700KF 16-Core CPU and an NVIDIA RTX 4070ti GPU.

\begin{figure}[t]
	\centering
	\includegraphics[width=1\linewidth]{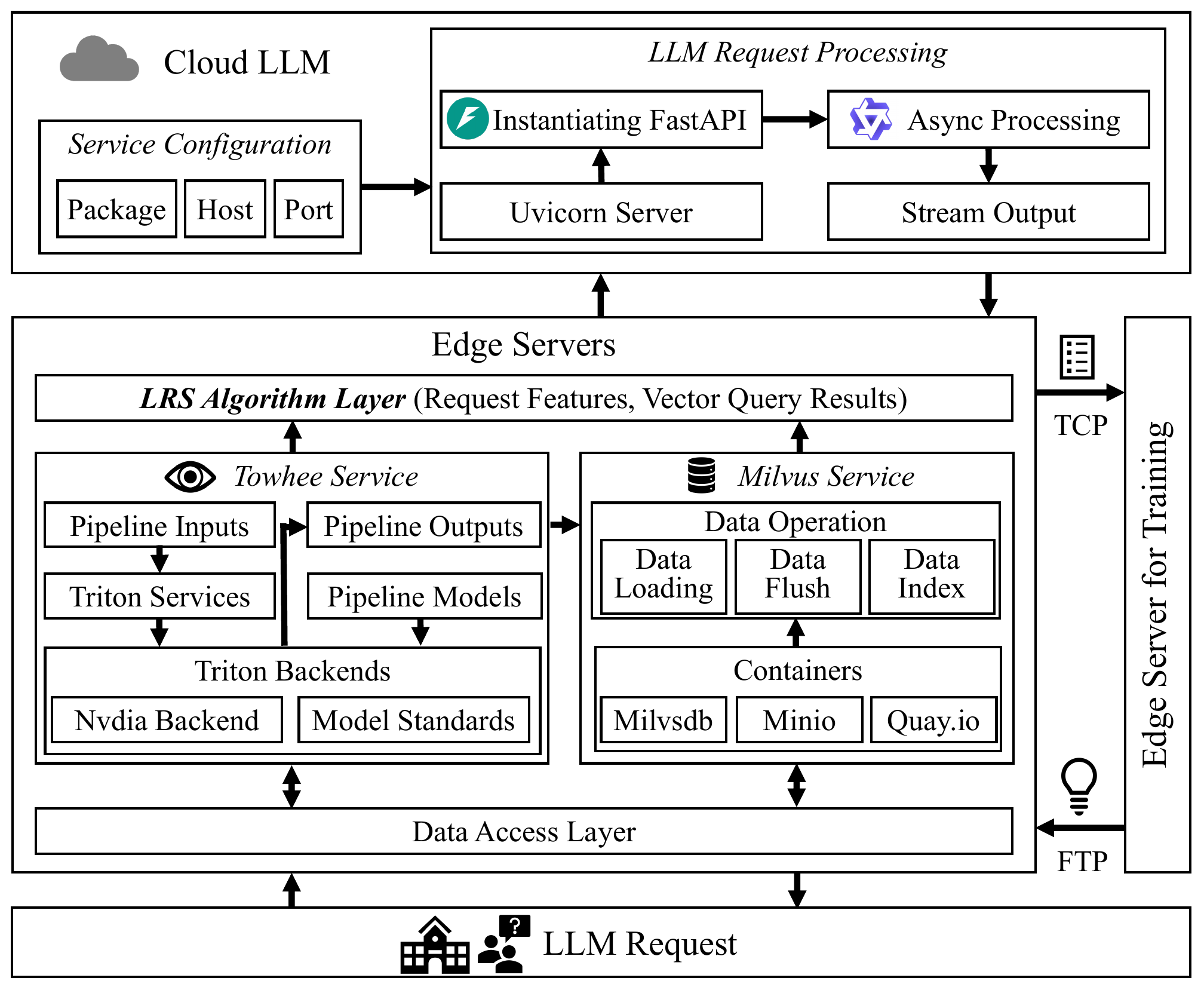}
	\centering
	\caption{System implementation details.}
	\label{fig:System overview}
\end{figure}

The Qwen7b quantized by int8 is deployed as the cloud LLM, based on FastAPI and Uvicorn servers \cite {b22}. A high-performance workstation with an Intel i9-14900KF 24-Core CPU, an NVIDIA RTX 4090 GPU, and 64GB RAM is used as the cloud. Given the slow processing speed for parallel LLM requests and the limited experimental devices, some experiments are conducted in virtual machines (VMs). Each VM has a minimum of four CPU cores, 50GB storage, and 8GB RAM. The system overview we have implemented is illustrated in Fig. \ref{fig:System overview}.

LLM requests from users can be offloaded to the nearest server and then vectorized through \textit{Towhee} service and input into a vector database to query relevant vectors. 
The \textit{LRS Algorithm Layer} can determine the scheduling location of LLM requests, including directly sending results to users through \textit{Action A} at the edge and sending LLM requests to the cloud through \textit{Action B} or \textit{Action C}. 
The cloud server outputs the answer to the LLM request after instantiating FastAPI.
Each server sends experience to the edge server for training through the TCP protocol and obtains the latest network weights through the FTP protocol.

\textbf{\textit{Milvus service:}} \textit{Milvus service} is an open-source vector database that enables vector similarity search \cite{b15}. \textit{Milvus} enables indexing and querying vector data through different containers. Vector data is usually stored on the hard drive through data flush and loaded into memory to accelerate queries as they are used. 
The type of collection index is \textit{IVF\_FLAT}, which is an index type based on inverted files. Based on this, we create 128 inverted lists for the inverted files. When performing a vector query, the 10 closest candidate items are searched from the inverted lists for precise distance calculation \cite{b15}.

\textbf{\textit{Towhee service:}} We use the open-source \textit{Towhee} framework and adopt \textit{gpt-neo-1.3B} as the embedding model\cite{b28}, \cite{b29}. This is a trained GPT-style language model, which supports multilingual embedding requests. We have deployed a \textit{Triton} acceleration framework based on the NVIDIA plugin on the server to improve embedding speed. Through the pipeline cached embedding model, LLM requests can be embedded in vectors with a dimension of 768. 

\subsection{Experimental Settings}

\textbf{Data preprocessing:} We use the multilingual open dialogue dataset \textit{oasst1} as LLM requests \cite{b20, b31}. Due to the multilingual and diverse nature of the LLM request, the highest-ranked dialogue from the \textit{oasst1} is chosen as the training set. Each item in the training set is expanded into five versions: English, Spanish, German, Chinese, and Russian using Qwen14b. The test set was obtained by restating the Question in QA pairs through Qwen14b.

\begin{figure*}[!h]
	\centering
	\includegraphics[width=7in]{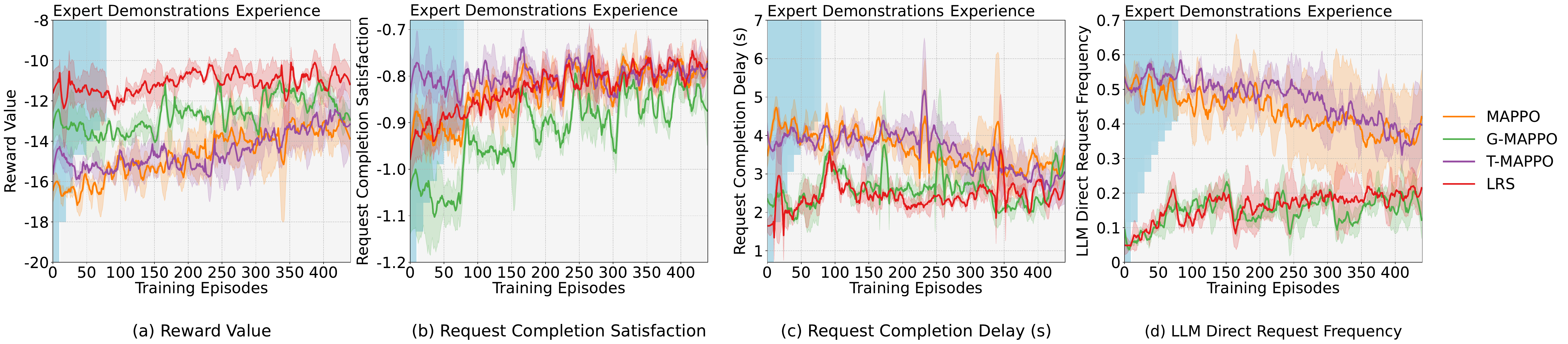}
	\centering
	\caption{Performance of LRS during training episodes with a weight ratio $\omega=0.1$}
	\label{fig:weight train}
\end{figure*}
\begin{figure*}[!h]
	\centering
	\includegraphics[width=7in]{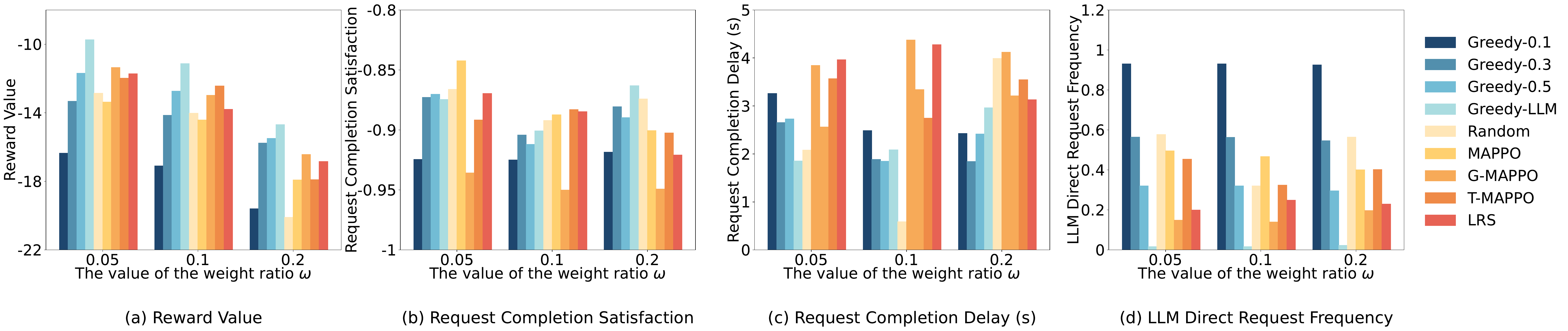}
	\centering
	\caption{Performance with different weight ratio $\omega$ when LLM requests are offloaded to the nearest edge server.}
	\label{fig:weight test}
\end{figure*}
\begin{figure*}[!h]
	\centering
	\includegraphics[width=7in]{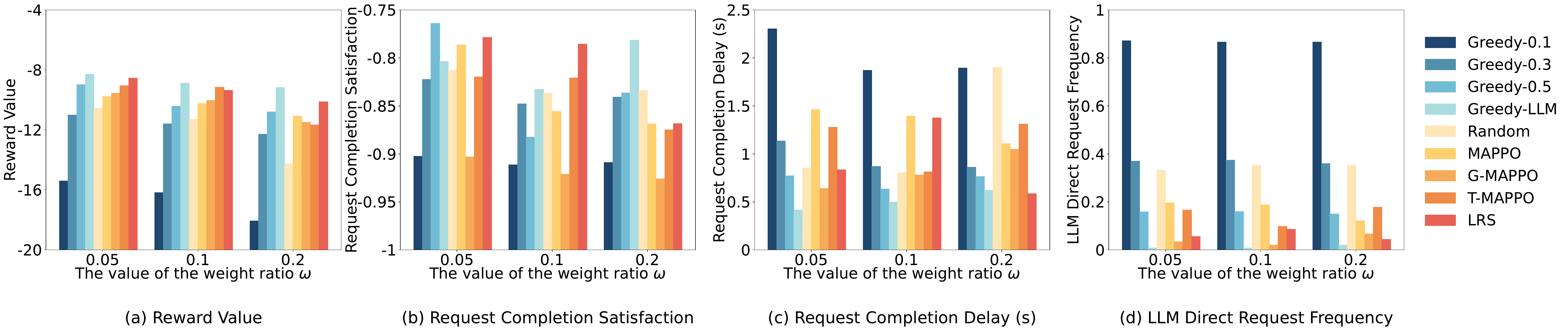}
	\centering
	\caption{Performance with different weight ratio $\omega$ when LLM requests are offloaded to all servers.}
	\label{fig:co weight test}
\end{figure*}

\textbf{Parameter settings:} For a training dataset of 3000 samples, 13500 rounds are used for training and 500 rounds for testing. As the dataset size increases, the number of training rounds is adjusted accordingly. Results are recorded every 300 rounds during the experiment. The learning rates for the policy and value networks are set to 0.0003 and 0.001, respectively. The discount factor is set to 0.99.
\begin{figure*}[!h]
	\centering
	\includegraphics[width=7in]{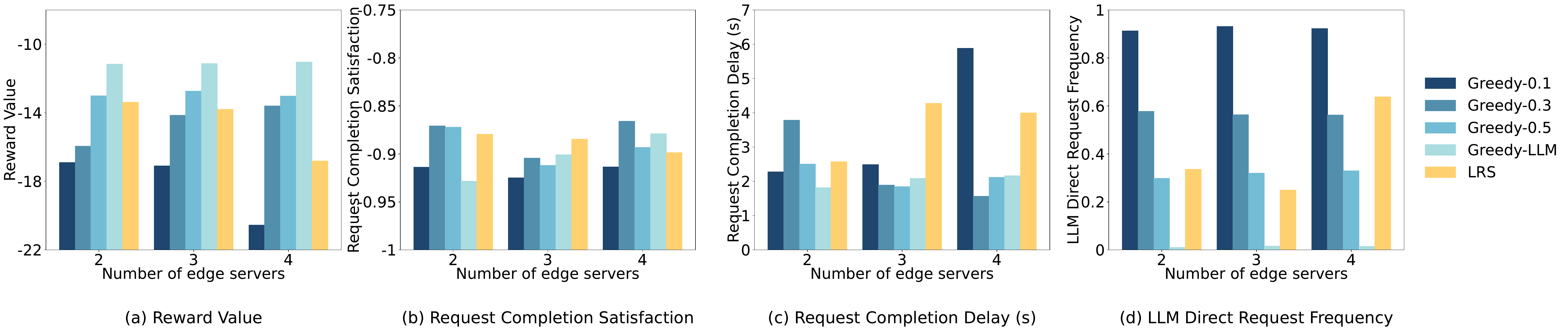}
	\centering
	\caption{Performance with different number of edge servers when LLM requests are offloaded to the nearest edge server.}
	\label{fig:nodes test}
\end{figure*}
\begin{figure*}[!h]
	\centering
	\includegraphics[width=7in]{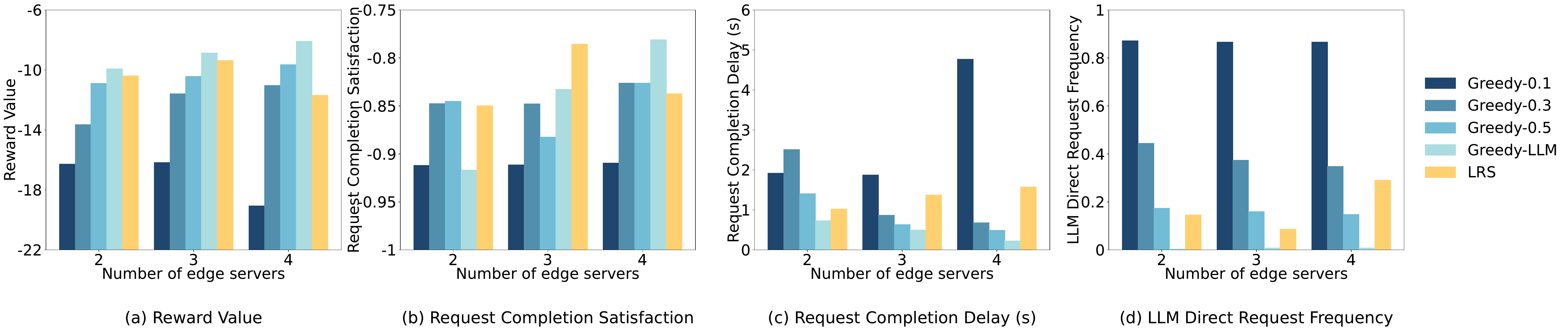}
	\centering
	\caption{Performance with different number of edge servers when LLM requests are offloaded to all servers.}
	\label{fig:co nodes test}
\end{figure*}
\begin{figure*}[!h]
	\centering
	\includegraphics[width=7in]{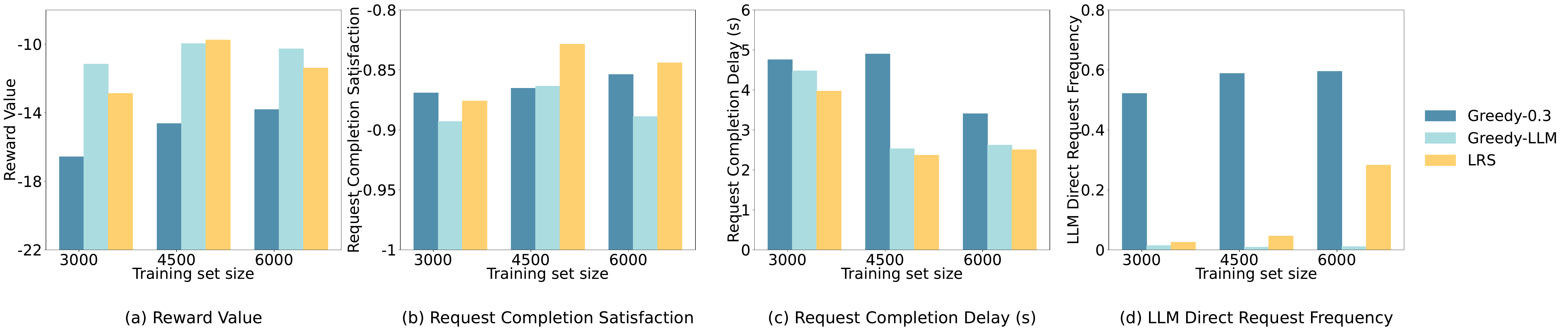}
	\centering
	\caption{Performance with different training set sizes when LLM requests are offloaded to the nearest edge server.}
	\label{fig:dataset test}
\end{figure*}
\begin{figure*}[!h]
	\centering
	\includegraphics[width=7in]{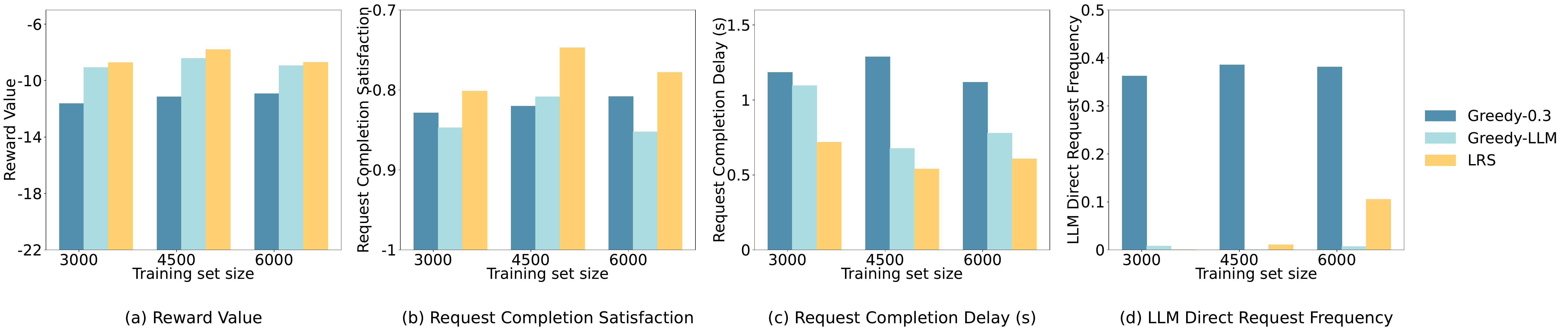}
	\centering
	\caption{Performance with different training set sizes when LLM requests are offloaded to all servers.}
	\label{fig:co dataset test}
\end{figure*}

\textbf{Baselines:} The following baselines are conducted.

\begin{enumerate}
	\item \textbf{Greedy-0.1, Greedy-0.3, Greedy-0.5}\cite{b32}\textbf{:} These algorithms make scheduling decisions by judging the vector query results and the fixed threshold. When the vector query results is higher than the threshold, the server directly requests the LLM; otherwise, it use the vector database to enhance the request. Experiments are conducted with thresholds of 0.1, 0.3, and 0.5.
	\item \textbf{Greedy-LLM:} The algorithm completes LLM requests with an initial estimated reward, updated periodically based on directly requesting LLM completion satisfaction. When the vector query result is better than the reward, the vector database is used to complete requests; otherwise, the server requests the LLM directly.
	\item \textbf{MAPPO}\cite{b30}\textbf{:} This algorithm is a multi-agent proximal policy optimization algorithm based on deep RL. 
	\item \textbf{G-MAPPO:} This algorithm is a MAPPO algorithm with expert demonstrations.
	\item \textbf{T-MAPPO:} This algorithm is a MAPPO algorithm with an external Transformer Encoder, which takes the connection of the extracted request vector features and the vector cache comparison results as input.
	\item  \textbf{Random:} Edge servers complete LLM requests by randomly selecting actions.
\end{enumerate}

\textbf{Evaluation methods:} Because of the differing quality of cached data for each server during training, we suggest two evaluation methods, which can be outlined as follows:

\begin{itemize}
	\item LLM requests are offloaded to the nearest server.
	\item LLM requests are offloaded to all servers, and the fastest response is returned as the answer to the request. While distributed servers collaborate in training, disparities in their processing capabilities arise from variations in vector databases.
\end{itemize}

\subsection{Experimental Results}
To showcase the effectiveness of the LRS algorithm, we carry out training and evaluation within the system. After every 300 LLM requests are fulfilled, we document an average outcome in the figures. Within the figures, LLM Direct Request Frequency represents the proportion of direct processing requests via LLM out of the total processed requests. The weight ratio is computed as $ w = \varphi _{2} / \varphi _{1}$. The reward value in the figures represents the average value achieved by all servers that complete LLM requests. To enhance the fitting ability of the network, we have increased the reward value in Eq. (\ref{formula-rt}) by a factor of 10.


\textbf{Performance with different reward weights:} Fig. \ref{fig:weight train} illustrates the performance of the algorithms during training episodes with $ w = 0.1$. The shaded regions in Fig. \ref{fig:weight train} depict the proportional relationship between expert demonstrations and experience utilized in network training. The shaded segment of the curve in the figure reflects the mean and variance relationships of the evaluation results, indicating the performance and disparities of agents. From Fig. \ref{fig:weight train}(b) and Fig. \ref{fig:weight train}(c), it is evident that LRS's request completion satisfaction notably improves, while LLM's request completion delay remains relatively low. Furthermore, Fig. \ref{fig:weight train}(a) and Fig. \ref{fig:weight train}(d) reveal that LRS and G-MAPPO guided by expert demonstrations can attain higher initial reward values and a greater frequency of vector data reuse. Fig. \ref{fig:weight train} demonstrates that the shadow area of the LRS algorithm's curve is relatively small, indicating its superior ability to optimize multiple agents simultaneously. The overall stability of the LRS algorithm surpasses that of other algorithms.


From Fig. \ref{fig:weight test}(b) and Fig. \ref{fig:co weight test}(b), it is clear that LRS has a significant advantage with $w=0.1$. This is because changes in weight not only affect system performance but also network fitting. When $w$ is high, the rewards of all algorithms, including LRS and baseline algorithms, are relatively low, as shown in Fig. \ref{fig:weight test}(a). On the other hand, as shown in Fig. \ref{fig:weight test}(c), a low $w$ causes the system to prioritize delay less, resulting in a relatively high overall request completion delay. Therefore, we set $w$ to 0.1 in subsequent experiments.

From Fig. \ref{fig:weight test} and Fig. \ref{fig:co weight test}, the reward obtained with the Greedy-LLM algorithm is higher. This is due to the task scheduling strategy of the Greedy-LLM algorithm minimizing the use of Action C. While the algorithm decreases the latency of LLM request processing, it also raises the risk of incorrect matching between vector knowledge and LLM requests, leading to decreased satisfaction and reduced usability of the algorithm. This is evident from the experimental results in Fig. \ref{fig:co weight test}(b) and beyond.

When LLM requests are distributed to multiple servers, the request completion delay can be significantly reduced due to the varied capabilities of each server in handling LLM requests. With $w$ set at 0.1, LRS exhibits a reasonable LLM direct request frequency, enhancing the utilization of vector databases for collaboration with LLM and improving request completion quality. Despite not being optimal, LRS shows enhanced request completion satisfaction, along with a reduction in LLM direct request frequency for similar requests, as depicted in Fig. \ref{fig:co weight test}(c) and Fig. \ref{fig:co weight test}(d). 

The experimental results indicate that the LRS algorithm can boost the overall reward value by up to 14.59\% with $w=0.1$ and the offloading of LLM requests to all servers. Moreover, the request completion satisfaction of the system is increased by 13.83\%, 7.37\%, 10.98\%, 5.65\%, 6.08\%, 8.22\%, 14.72\%, and 4.31\% on average compared with Greedy-0.1, Greedy-0.3, Greedy-0.5, Greed-LLM, Random, MAPPO, G-MAPPO, and T-MAPPO algorithms, respectively.



\textbf{Performance with different number of servers:} We also evaluate the algorithms' performance under different numbers of servers, as depicted in Fig. \ref{fig:nodes test} and Fig. \ref{fig:co nodes test}. Previous experiments compared LRS with other MARL and random algorithms. Given LRS's superior performance, these algorithms are not revisited in this section. As shown in Fig. \ref{fig:co nodes test}(a) and Fig. \ref{fig:co nodes test}(b), LRS excels with fewer servers. The performance of LRS is impacted by an increase in the number of servers when the network structure remains constant.


Unlike Greedy-LLM, which minimally utilizes LLM, LRS makes more judicious scheduling decisions, as depicted in Fig. \ref{fig:nodes test}(d) and Fig. \ref{fig:co nodes test}(d). The figures illustrate that LRS enhances the total reward value by up to 16.49\% when three servers are present in the system and the LLM requests are distributed across all servers. 

\textbf{Performance with different training set sizes:} After obtaining parameters with performance advantages, we further verify the performance of algorithms on different training set sizes. Fig. \ref{fig:dataset test} and Fig. \ref{fig:co dataset test} show the evaluation results on different training set sizes. As the training set size expands, the advantage of LRS becomes more prominent, indicating that LRS makes LLM request scheduling decisions with high QoS under optimal hyperparameter conditions. Furthermore, the LLM request frequency is not strictly inversely proportional to the QoS of request completion. The results show that the average reward value increases by 25.10\% and 4.59\%, respectively, compared with the Greedy-0.3 and Greed-LLM algorithms. Therefore, the LRS algorithm based on cloud-edge collaboration is a powerful way to improve QoS.

\section{Conclusion}
\label{sec5}

In this paper, we presented a framework for optimizing QoS in LLMs through a collaborative cloud-edge approach assisted by vector databases. We comprehensively modeled the LRS problem, considering both the satisfaction of the LLM request and the completion delay. We proposed a method for extracting LLM request features based on the Transformer Encoder and combined these features with the query result features of LLM requests to fully describe the request properties and their relevance to local vector data. Additionally, we introduced training and updating algorithms based on expert demonstrations to optimize the sparse LLM request features and address policy exploration challenges. Finally, we presented the LRS algorithm to enhance the QoS of LLM requests through cloud-edge collaboration. The system was deployed in a physical environment and evaluated using open-source QA datasets. Experimental results demonstrated that the LRS algorithm outperformed baseline algorithms by up to 15.31\%. Future work will focus on deploying and testing joint optimization problems that consider server resource usage and data characteristics. This will further enhance the efficiency and scalability of the proposed VELO framework in real-world applications.

\bibliographystyle{IEEEtran}
\bibliography{egbib}

\begin{thebibliography}{10}
\providecommand{\url}[1]{#1}
\csname url@samestyle\endcsname
\providecommand{\newblock}{\relax}
\providecommand{\bibinfo}[2]{#2}
\providecommand{\BIBentrySTDinterwordspacing}{\spaceskip=0pt\relax}
\providecommand{\BIBentryALTinterwordstretchfactor}{4}
\providecommand{\BIBentryALTinterwordspacing}{\spaceskip=\fontdimen2\font plus
\BIBentryALTinterwordstretchfactor\fontdimen3\font minus \fontdimen4\font\relax}
\providecommand{\BIBforeignlanguage}[2]{{%
\expandafter\ifx\csname l@#1\endcsname\relax
\typeout{** WARNING: IEEEtran.bst: No hyphenation pattern has been}%
\typeout{** loaded for the language `#1'. Using the pattern for}%
\typeout{** the default language instead.}%
\else
\language=\csname l@#1\endcsname
\fi
#2}}
\providecommand{\BIBdecl}{\relax}
\BIBdecl

\bibitem{b1}
Y.~Shen, J.~Shao, X.~Zhang, Z.~Lin, H.~Pan, D.~Li, J.~Zhang, and K.~B. Letaief, ``Large language models empowered autonomous edge ai for connected intelligence,'' \emph{IEEE Communications Magazine}, 2024, doi: {10.1109/MCOM.001.2300550}.

\bibitem{b2}
M.~Xu, H.~Du, D.~Niyato, J.~Kang, Z.~Xiong, S.~Mao, Z.~Han, A.~Jamalipour, D.~I. Kim, X.~Shen \emph{et~al.}, ``Unleashing the power of edge-cloud generative ai in mobile networks: A survey of aigc services,'' \emph{IEEE Communications Surveys \& Tutorials}, 2024, doi: {10.1109/COMST.2024.3353265}.

\bibitem{9808158}
C.~Ding, Z.~Lu, F.~Juefei-Xu, V.~N. Boddeti, Y.~Li, and J.~Cao, ``Towards transmission-friendly and robust cnn models over cloud and device,'' \emph{IEEE Transactions on Mobile Computing}, vol.~22, no.~10, pp. 6176--6189, 2023.

\bibitem{new_b}
Y.-C. Wang, J.~Xue, C.~Wei, and C.~C.~J. Kuo, ``An overview on generative ai at scale with edge–cloud computing,'' \emph{IEEE Open Journal of the Communications Society}, vol.~4, pp. 2952--2971, 2023.

\bibitem{new_c1}
J.~Kim, J.~H. Lee, S.~Kim, J.~Park, K.~M. Yoo, S.~J. Kwon, and D.~Lee, ``Memory-efficient fine-tuning of compressed large language models via sub-4-bit integer quantization,'' in \emph{Proceedings of the Advances in Neural Information Processing Systems (NeurIPS)}, vol.~36.\hskip 1em plus 0.5em minus 0.4em\relax Curran Associates, Inc., 2024, pp. 36\,187--36\,207.

\bibitem{new_b3}
G.~Xiao, J.~Lin, M.~Seznec, H.~Wu, J.~Demouth, and S.~Han, ``Smoothquant: Accurate and efficient post-training quantization for large language models,'' in \emph{Proceedings of the International Conference on Machine Learning (ICMR)}.\hskip 1em plus 0.5em minus 0.4em\relax PMLR, 2023, pp. 38\,087--38\,099.

\bibitem{new_b4}
C.-Y. Hsieh, C.-L. Li, C.-K. YEH, H.~Nakhost, Y.~Fujii, A.~J. Ratner, R.~Krishna, C.-Y. Lee, and T.~Pfister, ``Distilling step-by-step! outperforming larger language models with less training data and smaller model sizes,'' in \emph{Proceedings of the Annual Meeting Of The Association For Computational Linguistics (ACL)}.\hskip 1em plus 0.5em minus 0.4em\relax Association for Computational Linguistics, 2023, pp. 8003--8017.

\bibitem{new_b5}
M.~Lin, L.~Cao, Y.~Zhang, L.~Shao, C.-W. Lin, and R.~Ji, ``Pruning networks with cross-layer ranking \& k-reciprocal nearest filters,'' \emph{IEEE transactions on neural networks and learning systems}, vol.~34, no.~11, pp. 9139--9148, 2023.

\bibitem{b6}
Y.~Chen, R.~Li, Z.~Zhao, C.~Peng, J.~Wu, E.~Hossain, and H.~Zhang, ``Netgpt: A native-ai network architecture beyond provisioning personalized generative services,'' \emph{arXiv preprint arXiv:2307.06148}, 2023.

\bibitem{b7}
X.~Huang, P.~Li, H.~Du, J.~Kang, D.~Niyato, D.~I. Kim, and Y.~Wu, ``Federated learning-empowered ai-generated content in wireless networks,'' \emph{IEEE Network}, 2024, doi: {10.1109/MNET.2024.3353377}.

\bibitem{b71}
H.~Wang, J.~Hong, and J.~Luo, ``Service matching based on group preference and service representation learning for edge caching,'' in \emph{Proceedings of the 2023 IEEE International Conference on Web Services (ICWS)}.\hskip 1em plus 0.5em minus 0.4em\relax IEEE, 2023, pp. 1--6.

\bibitem{b8}
Q.~Cao, P.~Khanna, N.~D. Lane, and A.~Balasubramanian, ``Mobivqa: Efficient on-device visual question answering,'' in \emph{Proceedings of the ACM on Interactive, Mobile, Wearable and Ubiquitous Technologies}, vol.~6, no.~2, 2022, pp. 1--23.

\bibitem{b9}
A.~Nematallah, C.~H. Park, and D.~Black-Schaffer, ``Exploring the latency sensitivity of cache replacement policies,'' \emph{IEEE Computer Architecture Letters}, vol.~22, no.~2, pp. 93--96, 2023.

\bibitem{b10}
Y.~Han, C.~Liu, and P.~Wang, ``A comprehensive survey on vector database: Storage and retrieval technique, challenge,'' \emph{arXiv preprint arXiv:2310.11703}, 2023.

\bibitem{b15}
J.~Wang, X.~Yi, R.~Guo, H.~Jin, P.~Xu, S.~Li, X.~Wang, X.~Guo, C.~Li, X.~Xu \emph{et~al.}, ``Milvus: A purpose-built vector data management system,'' in \emph{Proceedings of the 2021 International Conference on Management of Data (SIGMOD)}, 2021, pp. 2614--2627.

\bibitem{b11}
H.~Jiang, Q.~Wu, C.-Y. Lin, Y.~Yang, and L.~Qiu, ``{LLML}ingua: Compressing prompts for accelerated inference of large language models,'' in \emph{Proceedings of the 2023 Conference on Empirical Methods in Natural Language Processing}.\hskip 1em plus 0.5em minus 0.4em\relax Association for Computational Linguistics, 2023, pp. 13\,358--13\,376.

\bibitem{b12}
J.~J. Pan, J.~Wang, and G.~Li, ``Survey of vector database management systems,'' \emph{arXiv preprint arXiv:2310.14021}, 2023.

\bibitem{b14}
O.~Topsakal and T.~C. Akinci, ``Creating large language model applications utilizing langchain: A primer on developing llm apps fast,'' in \emph{Proceedings of the International Conference on Applied Engineering and Natural Sciences (ICAENS)}, vol.~1, no.~1, 2023, pp. 1050--1056.

\bibitem{b16}
D.~Driess, F.~Xia, M.~S.~M. Sajjadi, C.~Lynch, A.~Chowdhery, B.~Ichter, A.~Wahid, J.~Tompson, Q.~Vuong, T.~Yu, W.~Huang, Y.~Chebotar, P.~Sermanet, D.~Duckworth, S.~Levine, V.~Vanhoucke, K.~Hausman, M.~Toussaint, K.~Greff, A.~Zeng, I.~Mordatch, and P.~Florence, ``{P}a{LM}-e: An embodied multimodal language model,'' in \emph{Proceedings of the 40th International Conference on Machine Learning}, vol. 202.\hskip 1em plus 0.5em minus 0.4em\relax PMLR, 2023, pp. 8469--8488.

\bibitem{b17}
T.~Ahmed and P.~Devanbu, ``Better patching using llm prompting, via self-consistency,'' in \emph{Proceedings of the 38th IEEE/ACM International Conference on Automated Software Engineering (ASE)}.\hskip 1em plus 0.5em minus 0.4em\relax IEEE, 2023, pp. 1742--1746.

\bibitem{x3}
Z.~Tang, W.~Jia, X.~Zhou, W.~Yang, and Y.~You, ``Representation and reinforcement learning for task scheduling in edge computing,'' \emph{IEEE Transactions on Big Data}, vol.~8, no.~3, pp. 795--808, 2020.

\bibitem{b18}
J.~K. Gupta, M.~Egorov, and M.~Kochenderfer, ``Cooperative multi-agent control using deep reinforcement learning,'' in \emph{Proceedings of the Autonomous Agents and Multiagent Systems (AAMAS)}.\hskip 1em plus 0.5em minus 0.4em\relax Springer, 2017, pp. 66--83.

\bibitem{x4}
H.~Zhu, X.~Li, L.~Chen, and R.~Ruiz, ``Smart offloading computation-intensive \& delay-intensive tasks of real-time workflows in mobile edge computing,'' in \emph{Proceedings of the 2023 IEEE International Conference on Web Services (ICWS)}.\hskip 1em plus 0.5em minus 0.4em\relax IEEE, 2023, pp. 695--697.

\bibitem{b25}
A.~Vaswani, N.~Shazeer, N.~Parmar, J.~Uszkoreit, L.~Jones, A.~N. Gomez, {\L}.~Kaiser, and I.~Polosukhin, ``Attention is all you need,'' in \emph{Proceedings of the Advances in Neural Information Processing Systems (NeurIPS)}, vol.~30.\hskip 1em plus 0.5em minus 0.4em\relax Curran Associates, Inc., 2017, p. 5998–6008.

\bibitem{9954278}
Z.~Lu, C.~Ding, F.~Juefei-Xu, V.~N. Boddeti, S.~Wang, and Y.~Yang, ``Tformer: A transmission-friendly vit model for iot devices,'' \emph{IEEE Transactions on Parallel and Distributed Systems}, vol.~34, no.~2, pp. 598--610, 2023.

\bibitem{b23}
R.~Guo, X.~Luan, L.~Xiang, X.~Yan, X.~Yi, J.~Luo, Q.~Cheng, W.~Xu, J.~Luo, F.~Liu \emph{et~al.}, ``Manu: a cloud native vector database management system,'' \emph{Proceedings of the VLDB Endowment (PVLDB)}, vol.~15, no.~12, pp. 3548--3561, 2022.

\bibitem{x5}
L.~Huang, X.~Feng, C.~Zhang, L.~Qian, and Y.~Wu, ``Deep reinforcement learning-based joint task offloading and bandwidth allocation for multi-user mobile edge computing,'' \emph{Digital Communications and Networks}, vol.~5, no.~1, pp. 10--17, 2019.

\bibitem{b19}
C.~Gulcehre, T.~L. Paine, B.~Shahriari, M.~Denil, M.~Hoffman, H.~Soyer, R.~Tanburn, S.~Kapturowski, N.~Rabinowitz, D.~Williams, G.~Barth-Maron, Z.~Wang, N.~de~Freitas, and W.~Team, ``Making efficient use of demonstrations to solve hard exploration problems,'' in \emph{Proceedings of the International Conference on Learning Representations (ICLR)}, 2020.

\bibitem{b22}
J.~Bai, S.~Bai, Y.~Chu, Z.~Cui, K.~Dang, X.~Deng, Y.~Fan, W.~Ge, Y.~Han, F.~Huang \emph{et~al.}, ``Qwen technical report,'' \emph{arXiv preprint arXiv:2309.16609}, 2023.

\bibitem{b20}
A.~K{\"o}pf, Y.~Kilcher, D.~von R{\"u}tte, S.~Anagnostidis, Z.~R. Tam, K.~Stevens, A.~Barhoum, D.~Nguyen, O.~Stanley, R.~Nagyfi \emph{et~al.}, ``Openassistant conversations-democratizing large language model alignment,'' in \emph{Proceedings of the Advances in Neural Information Processing Systems (NeurIPS)}, vol.~36.\hskip 1em plus 0.5em minus 0.4em\relax Curran Associates, Inc., 2024, pp. 47\,669--47\,681.

\bibitem{b31}
\BIBentryALTinterwordspacing
oasst1. [Online]. Available: \url{https://huggingface.co/datasets/OpenAssistant/oasst1}
\BIBentrySTDinterwordspacing

\bibitem{bqos}
Z.~Ye, W.~Gao, Q.~Hu, P.~Sun, X.~Wang, Y.~Luo, T.~Zhang, and Y.~Wen, ``Deep learning workload scheduling in gpu datacenters: A survey,'' \emph{ACM Computing Surveys}, vol.~56, no.~6, pp. 1--38, 2024.

\bibitem{b13}
P.~Wu, J.~Li, L.~Shi, M.~Ding, K.~Cai, and F.~Yang, ``Dynamic content update for wireless edge caching via deep reinforcement learning,'' \emph{IEEE Communications Letters}, vol.~23, no.~10, pp. 1773--1777, 2019.

\bibitem{b24}
Z.~Han, H.~Tan, G.~Chen, R.~Wang, Y.~Chen, and F.~C. Lau, ``Dynamic virtual machine management via approximate markov decision process,'' in \emph{Proceedings of the 35th Annual IEEE International Conference on Computer Communications (INFOCOM)}.\hskip 1em plus 0.5em minus 0.4em\relax IEEE, 2016, pp. 1--9.

\bibitem{x6}
Z.~Tang, F.~Mou, J.~Lou, W.~Jia, Y.~Wu, and W.~Zhao, ``Multi-user layer-aware online container migration in edge-assisted vehicular networks,'' \emph{IEEE/ACM Transactions on Networking}, vol.~32, no.~2, pp. 1807--1822, 2024.

\bibitem{b28}
S.~Black, S.~Biderman, E.~Hallahan, Q.~Anthony, L.~Gao, L.~Golding, H.~He, C.~Leahy, K.~McDonell, J.~Phang, M.~Pieler, U.~S. Prashanth, S.~Purohit, L.~Reynolds, J.~Tow, B.~Wang, and S.~Weinbach, ``{GPT}-{N}eo{X}-20{B}: An open-source autoregressive language model,'' in \emph{Proceedings of BigScience Episode {\#}5 -- Workshop on Challenges {\&} Perspectives in Creating Large Language Models}.\hskip 1em plus 0.5em minus 0.4em\relax Association for Computational Linguistics, 2022, pp. 95--136.

\bibitem{b26}
R.~S. Sutton, D.~McAllester, S.~Singh, and Y.~Mansour, ``Policy gradient methods for reinforcement learning with function approximation,'' in \emph{Proceedings of the Advances in Neural Information Processing Systems (NeurIPS)}, vol.~12.\hskip 1em plus 0.5em minus 0.4em\relax MIT Press, 1999, p. 1057–1063.

\bibitem{b27}
J.~Schulman, F.~Wolski, P.~Dhariwal, A.~Radford, and O.~Klimov, ``Proximal policy optimization algorithms,'' \emph{arXiv preprint arXiv:1707.06347}, 2017.

\bibitem{b33}
I.~Sarkar, M.~Adhikari, S.~Kumar, and V.~G. Menon, ``Deep reinforcement learning for intelligent service provisioning in software-defined industrial fog networks,'' \emph{IEEE Internet of Things Journal}, vol.~9, no.~18, pp. 16\,953--16\,961, 2022.

\bibitem{b34}
K.~Han, A.~Xiao, E.~Wu, J.~Guo, C.~XU, and Y.~Wang, ``Transformer in transformer,'' in \emph{Proceedings of the Advances in Neural Information Processing Systems (NeurIPS)}, vol.~34.\hskip 1em plus 0.5em minus 0.4em\relax Curran Associates, Inc., 2021, pp. 15\,908--15\,919.

\bibitem{b29}
L.~Gao, S.~Biderman, S.~Black, L.~Golding, T.~Hoppe, C.~Foster, J.~Phang, H.~He, A.~Thite, N.~Nabeshima \emph{et~al.}, ``The pile: An 800gb dataset of diverse text for language modeling,'' \emph{arXiv preprint arXiv:2101.00027}, 2020.

\bibitem{b32}
\BIBentryALTinterwordspacing
Fastgpt. [Online]. Available: \url{https://github.com/labring/FastGPT}
\BIBentrySTDinterwordspacing

\bibitem{b30}
C.~Yu, A.~Velu, E.~Vinitsky, J.~Gao, Y.~Wang, A.~Bayen, and Y.~Wu, ``The surprising effectiveness of ppo in cooperative multi-agent games,'' in \emph{Proceedings of the Advances in Neural Information Processing Systems (NeurIPS)}, vol.~35.\hskip 1em plus 0.5em minus 0.4em\relax Curran Associates, Inc., 2022, pp. 24\,611--24\,624.

\end{thebibliography}

\end{document}